\documentclass[letterpaper, 10 pt, conference]{ieeeconf}  

\IEEEoverridecommandlockouts                              

\usepackage{times}

\usepackage{tabularx}
\usepackage{booktabs}
\usepackage{graphicx}
\usepackage{multirow}
\usepackage{censor}
\usepackage{amssymb}
\usepackage{array}
\usepackage{amsmath}
\usepackage{cleveref}
\usepackage{cite}
\crefname{figure}{Fig.}{Figs.}
\Crefname{figure}{Fig.}{Figs.}

\crefname{table}{Table}{Tables}
\Crefname{table}{Table}{Tables}

\crefname{section}{Sec.}{Secs.}
\Crefname{section}{Sec.}{Secs.}

\newcolumntype{Y}{>{\centering\arraybackslash}X}

\title{\LARGE \bf
StageGuard: Learning Stage Transitions for Long-Horizon Robot Tasks via Agentic Distillation
}

\author{
Jinbang Huang$^{1}$,
Yuanzhao Hu$^{2,*}$,
Zhiyuan Li$^{3,*}$,
Ran Qi$^{3,*}$,
Yixin Xiao$^{1}$,
Yangzheng Wu$^{1}$,\\
Tengyue Ba$^{5}$,
Zhanguang Zhang$^{1}$,
Yingxue Zhang$^{1}$%
\thanks{$^{1}$Huawei Noah's Ark Lab,
$^{2}$University of British Columbia,
$^{3}$University of Toronto,
$^{4}$McGill University,
$^{5}$Department of Foundation Model, 2012 Labs.
$^{*}$Work done during an internship at Huawei Noah's Ark Lab. Corresponding to: \texttt{jinbang.huang.work@gmail.com,\{zhanguang.zhang, yingxue.zhang\}@huawei.com.}}%
}

\begin{document}

\maketitle
\thispagestyle{empty}
\pagestyle{empty}

\begin{abstract}

Hierarchical planning frameworks combine skills from multiple robot control policies for long-horizon task execution, where determining when to terminate the current skill and advance to the next subtask is essential. Existing approaches often rely on pre-designed completion signal checkers that are hard to obtain in real-world execution. Large-scale vision-language models (VLMs) offer strong reasoning capabilities, but their decision boundaries are not inherently aligned with task completion criteria, while cloud deployment and lengthy reasoning introduce substantial latency, limiting real-time monitoring. We propose StageGuard, an agentic distillation framework for accurate and efficient stage-transition decisions. StageGuard combines teacher-model reasoning with demonstration trajectories to generate structured explanations of subtask completion and policy switching. A lightweight student VLM uses these explanations to generate compact self-explanations, which are used for supervised fine-tuning. We evaluate stage-transition prediction on trajectories from two benchmarks and assess closed-loop task success through integration into hierarchical robot control on BEHAVIOR-1K, with further validation on real robots. Results show substantial improvements in stage-transition prediction while supporting efficient online monitoring.

\end{abstract}


\section{Introduction}
\label{sec:introduction}

Long-horizon robotic tasks require coordinated skills that establish the conditions for subsequent execution. Hierarchical planning frameworks compose these skills by decomposing goals into subtasks and assigning them to robot control policies~\cite{Kaelbling2011-mz,Silver2021-mv,Dalal2023-cq,mao2023learning,ye2026uniplan}. Recent work extends this paradigm to agentic robotic systems~\cite{lu2026aspire,xiao2026enpire}, some of which orchestrate heterogeneous policies to accomplish complex tasks~\cite{huang2026roboharness}.
Thus, a critical challenge also lies in determining when to terminate the current skill and advance to the next subtask. Incorrect stage transitions can compromise execution even when the task plan and individual policies are effective.
Stage-transition decisions require interpreting observations alongside execution history and intended subtask outcomes~\cite{10610216}. Premature switching leaves required conditions unsatisfied, while delayed switching can cause system latency or disturb an achieved state; both errors can propagate through subsequent subtasks. Hierarchical planning frameworks often rely on predefined skill termination conditions\cite{Liang2024-hf,han2024interpret,Silver2023-mi,Huang2025-ue,} or privileged simulator signals\cite{liu2023libero,han2025robocerebra}. In real-world execution, however, completion should be inferred from observations and task context, and termination conditions are difficult to specify for learned policies\cite{black2024,kim2024openvla,zheng2026xvla,li2025unified, kim2026cosmos}.

Large-scale vision-language models (VLMs) offer a promising foundation through their visual understanding and reasoning capabilities\cite{gemmateam2024,mu2023embodiedgpt,driess2023,team2025gemini}, but direct deployment of large-scale VLMs as execution monitors faces two challenges. First, their decision boundaries are not inherently aligned with subtask completion criteria, so plausible scene interpretations may yield incorrect transition judgments. Second, lengthy reasoning and communication delays in cloud-hosted deployment increase latency, limiting monitoring frequency. Demonstration trajectories indicate when subtasks are completed, but transition labels alone do not explain the visual evidence or task dependencies underlying each decision~\cite{Huang2025-zo,zhao2025cot}. Thus, explicitly linking observations and task context to the decision boundaries governing completion and policy switching is essential.

We propose \textbf{StageGuard}, an agentic distillation framework for learning accurate and efficient stage-transition decisions. StageGuard uses a teacher model within an agentic workflow to generate structured reasoning traces that explicitly link observations to ground-truth transition decisions in demonstration trajectories. Through learning by explanation, a lightweight student VLM acquires these decision patterns while reducing inference costs, serving as an efficient execution monitor for hierarchical robot control.
We evaluate StageGuard on LIBERO~\cite{liu2023libero}, BEHAVIOR-1K~\cite{li2024behavior1k}, and a real robot system to assess its impact on stage transition success. Results demonstrate improved stage-transition prediction and low inference latency.
Our main contributions are:
(1) A demonstration-grounded agentic distillation framework that constructs multi-layer teacher explanations connecting visual evidence and task requirements to annotated stage-transition decisions.
(2) A learning-by-explanation pipeline that reformulates teacher reasoning into compact student self-explanations and uses them for supervised fine-tuning of a lightweight execution monitor.
(3) Evaluations on LIBERO and BEHAVIOR-1K demonstrates improved stage-transition prediction and efficient inference. Closed-loop experiments in simulation and on real robots show improved task success.

\begin{figure*}[t]
    \centering
    \includegraphics[width=\textwidth]{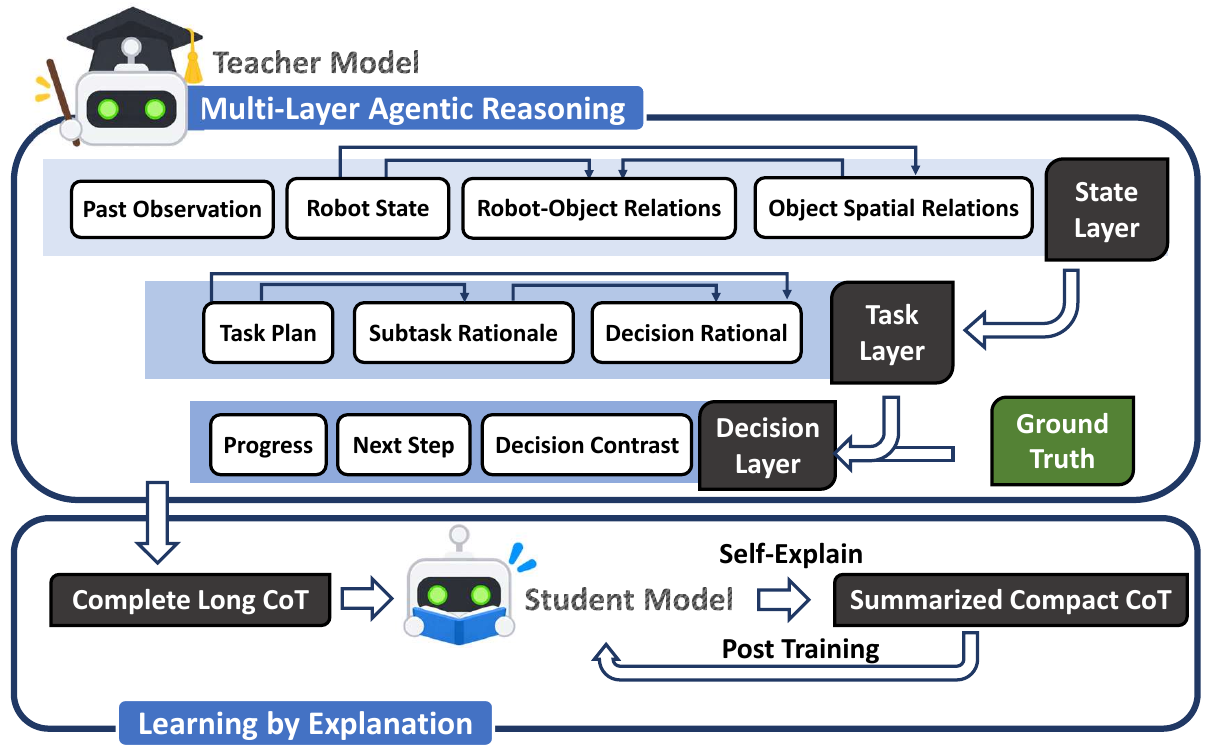}
    \caption{\textbf{Overview of StageGuard.}
    \textbf{Top:} A teacher model generates structured reasoning traces
    from demonstration trajectories using multi-layer agentic reasoning.
    The \textbf{state layer} analyzes observation history, robot state,
    robot--object relations, and object spatial relations.
    The \textbf{task layer} connects the task plan with subtask and
    decision rationales.
    The \textbf{decision layer} assesses task progress, identifies the
    next step, and determines stage transitions, guided by demonstration
    ground truth.
    \textbf{Bottom:} Through learning by explanation, a student VLM
    uses complete teacher-generated chain-of-thought (CoT) traces
    as context to generate compact self-explanations.
    Teacher-selected explanations and their corresponding monitoring
    targets are then used for supervised fine-tuning, enabling accurate
    and efficient subtask-completion and policy-switching decisions.}
    \label{fig:stageguard_overview}
\end{figure*}

\section{Related Work}

\subsection{Hierarchical Planning}

Hierarchical planning coordinates long-horizon robot tasks by decomposing goals into executable subtasks. Task and motion planning (TAMP) integrates symbolic action planning with continuous motion feasibility, using action preconditions and effects to organize successive operations~\cite{Dalal2023-cq,Huang2024-it,pmlr-v229-kumar23a,Silver2023-mi,Liang2024-hf}. Agentic robotic systems further expand how skills are developed and coordinated for self-evolution~\cite{lu2026aspire,xiao2026enpire}, model regulation~\cite{zhang2026harness}, and heterogeneous policy orchestration~\cite{huang2026roboharness}. These developments make execution monitoring an important component of hierarchical control: the system must determine whether the current skill has achieved its intended outcome before advancing~\cite{lee2021adversarial,lee2019composing}. Premature or delayed stage transitions can compromise the reliability of long-horizon task execution. Although large-scale VLMs can assess subtask completion, their inference latency constrains the frequency of online monitoring. StageGuard supports hierarchical planning and policy orchestration through a lightweight execution monitor that combines high stage-transition accuracy with efficient inference.

\subsection{Robot Progress Learning}

Visual progress models learn functions that quantify advancement toward task goals, providing feedback for policy learning and execution assessment~\cite{ma2023liv,alakuijala2025videolanguage,11445032,intelligence2025pi,wu2026large}. The SARM series further incorporates task stages or action primitives into progress estimation for long-horizon manipulation~\cite{chen2026sarm,chen2026sarm2}, while multiview approaches extend progress prediction to mobile manipulation trajectories~\cite{zoppellari2026multiview}. These methods primarily optimize progress or reward estimation, rather than directly learning when subtask completion conditions warrant a transition to the next policy.

VLM-based reasoning approaches support task planning, success assessment, and failure explanation~\cite{10610216,duan2024aha}, offering capabilities complementary to StageGuard's focus on learning subtask boundaries and switching times. Another line of work elicits progress and completion judgments through prompting and in-context reasoning~\cite{ma2024generative,schroeder2026rover}. Without task-specific adaptation, prompting-based methods inherit the backbone's accuracy-latency trade-off: larger VLMs are typically slower, while smaller models may yield unreliable state-transition judgments. StageGuard achieves both accuracy and efficiency through agentic distillation, aligning a lightweight monitor with demonstrated subtask boundaries and transition timing.

\section{Preliminaries}

\subsection{Task setting and demonstration data}
We consider a robot executing a high-level goal \(g\) according to a task plan \(\mathcal P\), which specifies subtasks and their intended execution order. At monitoring round \(t\), let \(u_t\) denote the current task specification in natural language, \(o_t\) the current observation, and \(h_{t-1}\) the semantic history summarizing preceding observations and subtask execution events, with \(h_0=\emptyset\).

The monitoring decision \(d_t\) belongs to the decision space
\begin{equation}
\mathcal A =
\{\mathtt{continue},\mathtt{advance},\mathtt{skip}\}.
\end{equation}
Here, \texttt{continue} retains the current subtask; \texttt{advance} indicates that its completion conditions are satisfied; and \texttt{skip} indicates an unrecoverable failure requiring the current subtask to be bypassed. The structured monitoring output is \(y_t=(d_t,v_t,h_t)\), where \(v_t\) denotes the next task specification, when applicable, and \(h_t\) is the updated semantic history.

The demonstration dataset is
\(\mathcal D = \{(g_i,\mathcal P_i,\tau_i)\}_{i=1}^{N}\),
where \(N\) is the number of demonstrations, and each demonstration contains a goal \(g_i\), a task plan \(\mathcal P_i\), and an annotated trajectory
\begin{equation}
\tau_i =
\bigl[(o_{i,t},u_{i,t},d_{i,t}^{\star},v_{i,t}^{\star})\bigr]_{t=1}^{T_i}.
\end{equation}
Here, \(T_i\) is the number of annotated monitoring rounds, and the superscript \(\star\) denotes ground-truth annotations of the decision and next task.

\subsection{Teacher and student models}
Let \(M_{\mathrm T}\) denote the teacher VLM and \(M_\theta\) the lightweight student VLM with parameters \(\theta\). The monitoring input is
\begin{equation}
x_t=(g,u_t,o_t,h_{t-1},\mathcal P).
\end{equation}
The teacher \(M_{\mathrm T}\) produces a reasoning trace \(r_t^{\mathrm T}\) for \(x_t\), which the initial student \(M_{\theta}\) condenses into a compact trace \(r_t^{\mathrm S}\). The resulting training dataset is
\begin{equation}
\mathcal C =
\bigl\{(x_{i,t},r_{i,t}^{\mathrm S},y_{i,t})\bigr\}_{i,t},
\end{equation}
where \(i\) and \(t\) index demonstrations and monitoring rounds, respectively. Let \(M_\theta^{\mathrm{FT}}\) denote the student obtained by fine-tuning \(M_{\theta}\) on \(\mathcal C\). It serves as the efficient execution monitor:
\begin{equation}
(r_t^{\mathrm S},y_t)=M_\theta^{\mathrm{FT}}(x_t).
\end{equation}

\section{Problem Formulation}

Given the demonstration dataset \(\mathcal D\), teacher VLM \(M_{\mathrm T}\), and student VLM \(M_\theta\), we seek to obtain an accurate and efficient execution monitor \(M_\theta^{\mathrm{FT}}\) by fine-tuning \(M_\theta\) on \(\mathcal C\). This corpus pairs structured monitoring outputs \(y_t\) with compact student explanations \(r_t^{\mathrm S}\) condensed from teacher reasoning \(r_t^{\mathrm T}\). Our objective is to learn monitoring decisions \(d_t\) and next-task predictions \(v_t\) aligned with demonstrated subtask boundaries and transition timing while maintaining low inference latency. During execution, \(M_\theta^{\mathrm{FT}}(x_t)\) produces \((r_t^{\mathrm S},y_t)\), with \(h_t\) included in \(x_{t+1}\).

\section{Methodology}

StageGuard distills demonstration-grounded reasoning into a lightweight execution monitor, as indicated in \Cref{fig:stageguard_overview}.
Multi-layer agentic reasoning generates teacher explanations that connect observed execution states with annotated subtask boundaries.
Learning by explanation converts these explanations into verified compact Chain-of-Thought(CoT) traces for student fine-tuning.

\subsection{Multi-Layer Agentic Reasoning}

We organize teacher reasoning into three layers: state, task, and decision.
Each layer comprises specialized agents instantiated through role-specific prompts to \(M_{\mathrm T}\), with their outputs forming structured reasoning of corresponding aspects.
Together, they construct \(r_t^{\mathrm T}\) by connecting observable evidence, task requirements, and demonstrated monitoring decisions.

\paragraph{State layer.}
The state layer characterizes the execution context through four agents with explicit information dependencies.
The \textbf{past-observation agent} extracts relevant changes and preceding execution events from the semantic history \(h_{t-1}\).
The \textbf{robot-state agent} examines \(o_t\) to characterize the observable robot configuration and gripper status.
Its output is passed to the \textbf{object spatial relation agent}, which uses this context together with \(o_t\) to examine relations among objects and target regions.
The \textbf{robot--object relation agent} then combines the robot-state and object-spatial analyses to identify interactions between the robot and task-relevant objects, including grasping, holding, and releasing.
The four agents' outputs are aggregated and passed to the task layer, providing both historical context and current physical evidence.

\paragraph{Task layer.}
Conditioned on the state-layer outputs, the task layer independently derives a monitoring decision and its rationale without access to ground-truth annotations.
The \textbf{task-plan agent} locates the current subtask within \(\mathcal P\) and identifies its dependencies and intended successors.
Its analysis is passed to the \textbf{subtask-rationale agent}, which explains how the current subtask contributes to \(g\) and specifies its completion conditions.
The \textbf{decision-rationale agent} combines both analyses with the state-layer evidence to infer whether to continue, advance, or skip, explaining the evidence and conditions supporting its judgment.

\paragraph{Decision layer.}
The decision layer evaluates task-layer reasoning using ground-truth annotations.
The \textbf{progress agent} receives auxiliary supervision by binning the ratio of completed to total control frames within the annotated subtask segment into semantic progress stages: \texttt{starting}, \texttt{early}, \texttt{middle}, \texttt{later}, and \texttt{completing}.
The \textbf{next-step agent} checks the intended successor from the task-plan analysis against \(v_t^\star\), when applicable, and explains any discrepancy.
The \textbf{decision contrast agent} compares the initial monitoring decision with \(d_t^\star\), explaining the supporting evidence when they agree, or identifying the mistaken inference and justifying the annotated decision when they differ.
The three layers' analyses and explanations form \(r_t^{\mathrm T}\), paired with the structured target \(y_t=(d_t^\star,v_t^\star,h_t)\), where \(h_t\) summarizes the updated execution context.
Ground-truth annotations are used during corpus construction only.

\subsection{Learning by Explanation}

Directly training a lightweight student on complete teacher CoT presents two difficulties.
Lengthy traces increase training and inference costs, while the teacher's reasoning structure and level of detail may exceed what the student can reliably reproduce, creating a gap between teacher supervision and the student's own reasoning patterns.
Our ablations support this concern: full-CoT training performs reasonably on simple tasks but is less effective on challenging ones.
We therefore use student self-explanation to construct compact reasoning supervision, followed by quality monitoring and supervised fine-tuning.

\paragraph{Student self-explanation.}
Given \(x_t\), the teacher trace \(r_t^{\mathrm T}\), and the structured target \(y_t\), we prompt the initial student \(M_{\theta}\) to explain the monitoring decision in its own words.
The student produces a compact trace \(r_t^{\mathrm S}\) that retains the decisive observation evidence, relevant completion or failure conditions, and justification for the monitoring decision and next task.
This process uses teacher reasoning as explanatory context while allowing the student to reformulate it into reasoning patterns that it can more readily express and learn.

\paragraph{Compact CoT monitoring.}
To control the quality of student-generated explanations, we generate multiple candidates through parallel student inference and use \(M_{\mathrm T}\) as judge to select the compact trace \(r_t^{\mathrm S}\).
The judge evaluates whether each candidate is grounded in \(x_t\), preserves the decision-relevant reasoning in \(r_t^{\mathrm T}\), and consistently justifies \(d_t^\star\) and \(v_t^\star\).
Selection favors explanations that retain the evidence needed to support the decision while avoiding unnecessary detail.
The selected explanations and their structured targets form the training corpus \(\mathcal C\).

\paragraph{Weighted supervised fine-tuning.}
We fine-tune \(M_\theta\) on \(\mathcal C\) to jointly generate the compact explanation and structured monitoring output.
Let \(z=(r^{\mathrm S},y)\) denote their serialized concatenation, with \(z_\ell\) its \(\ell\)-th token.
We use a weighted autoregressive language modeling objective:
\begin{equation}
\mathcal L_{\mathrm{SFT}}(\theta)
=
-\mathbb E_{(x,r^{\mathrm S},y)\sim\mathcal C}
\left[
\sum_{\ell=1}^{|z|}
w_\ell
\log p_\theta(z_\ell\mid x,z_{<\ell})
\right],
\end{equation}
where
\begin{equation}
w_\ell =
\begin{cases}
\lambda, & z_\ell \text{ belongs to } r^{\mathrm S},\\
1, & z_\ell \text{ belongs to } y.
\end{cases}
\end{equation}
The hyperparameter \(\lambda>0\) controls the relative contribution of explanation learning and structured-output supervision.
The resulting monitor \(M_\theta^{\mathrm{FT}}\) independently generates compact reasoning and monitoring outputs directly from \(x_t\), carrying the updated history \(h_t\) into the next monitoring round.
Teacher reasoning and parallel quality monitoring are required only during training-data construction.

\begin{table*}[t]
\centering
\caption{Stage-transition evaluation and inference efficiency on LIBERO.
The first three metrics are reported as percentages, and inference
efficiency is measured in Hz. Higher values are better for all metrics
($\uparrow$). The best and second-best distinct values are shown in
\textbf{bold} and \underline{underlined}, respectively; ties receive
the same formatting.}
\label{tab:stageguard_libero}

\fontsize{11pt}{11pt}\selectfont
\setlength{\tabcolsep}{3pt}
\renewcommand{\arraystretch}{1.05}

\begin{tabular*}{\textwidth}{@{\extracolsep{\fill}}lcccc@{}}
\toprule
Method
& \shortstack{Transition\\completion}
& \shortstack{Full\\trajectories}
& \shortstack{Next-subtask\\accuracy}
& \shortstack{Inference\\speed} \\
\midrule

GPT-5.6 Luna
& 37.92 & 39.50 & 20.94 & 0.29 \\

Gemma 4 E4B
& 0.00 & 0.00 & 0.00 & 3.55 \\

Gemma 4 12B Unified
& 1.51 & 9.50 & 0.19 & 3.05 \\

Qwen3.5-4B
& 17.36 & 22.75 & 5.47 & 0.62 \\

Qwen3.5-0.8B
& 0.00 & 0.00 & 0.00 & 2.67 \\

\midrule

ROVER-4B
& 26.23 & 30.50 & 5.47 & 0.74 \\

ROVER-0.8B
& 0.00 & 0.00 & 0.00 & \underline{3.64} \\

\midrule

StageGuard (w/o explain)
& \textbf{96.23}
& \textbf{98.00}
& \underline{74.15}
& 0.82 \\

StageGuard (Decision-Only)
& \underline{94.34}
& 92.50
& 69.43
& \textbf{47.60} \\

\textbf{StageGuard (ours)}
& \textbf{96.23}
& \underline{97.50}
& \textbf{89.81}
& 1.03 \\

\bottomrule
\end{tabular*}
\end{table*}

\section{Experiment Implementation}

\subsection{Training Data and Benchmarks}

We train and evaluate StageGuard on two datasets: LIBERO-Logic~\cite{huang2026hwm} and BEHAVIOR-1K~\cite{li2024behavior1k}, covering tabletop and mobile manipulation with different task complexities and execution horizons. LIBERO-Logic augments the standard LIBERO dataset~\cite{liu2023libero} with ground-truth subtask segmentation and state-transition annotations. Its relatively simple, short-horizon tabletop manipulation tasks enable the study of stage-transition monitoring in constrained workspaces. BEHAVIOR-1K provides household mobile manipulation tasks with ground-truth subtask segmentation and state-transition annotations. Due to computational constraints, we restrict our evaluation to six tasks: turning on radio, picking up trash, moving boxes to storage, installing a modem, collecting aluminum cans, and installing a scanner. The provided annotations support demonstration-grounded teacher reasoning generation and stage-transition evaluation. These tasks combine navigation, object search, and manipulation over longer execution horizons, with substantial viewpoint and background changes that challenge many VLMs. For each task in both benchmarks, we use 10 trajectories for training and use the remaining trajectories for evaluation.

\subsection{Implementation}

We use Qwen3.5-397B-A17B~\cite{team2026qwen3} as the teacher VLM \(M_{\mathrm T}\) and Qwen3.5-0.8B~\cite{team2026qwen3} as the lightweight student VLM \(M_\theta\).
For each training sample, we generate three compact CoT candidates through parallel student inference and use the teacher model as a judge to select one for supervision.
We fine-tune all parameters of the student on \(\mathcal C\) with a batch size of \(8\), using the weighted autoregressive objective with \(\lambda=0.5\) chosen empirically.
This assigns a weight of \(0.5\) to compact reasoning tokens and \(1.0\) to structured-output tokens, placing greater emphasis on accurate monitoring outputs.
On LIBERO-Logic, all fine-tuning runs use a learning rate of \(10^{-5}\) for \(10{,}000\) steps.
On BEHAVIOR-1K, training runs for \(35{,}000\) steps.

\subsection{Baselines}

We evaluate multiple VLM baselines and VLM-based progress reasoning methods for stage-transition monitoring.
The model baselines include the open-weight models Gemma 4 E4B and Gemma 4 12B Unified~\cite{team2026gemma}, Qwen3.5-4B and Qwen3.5-0.8B~\cite{team2026qwen3}, and the proprietary model GPT-5.6 Luna~\cite{openai2026gpt56}.
Given the monitoring input \(x_t\), each model is prompted to produce the structured output \(y_t=(d_t,v_t,h_t)\), comprising the monitoring decision, next-task specification, and updated semantic history.
We additionally adapt ROVER~\cite{schroeder2026rover}, which uses recursive subtask reasoning and a sliding visual context for progress estimation, to predict stage-transition decisions \(d_t\in\mathcal A\) and next-task specifications \(v_t\).
We instantiate the adapted framework with Qwen3.5-0.8B and Qwen3.5-4B, denoted as \textbf{ROVER-0.8B} and \textbf{ROVER-4B}, respectively.

\subsection{Ablation Study Setting}
We evaluate two variants to examine the contributions of learning by explanation and reasoning supervision.
\textbf{StageGuard (w/o explain)} directly fine-tunes the student on teacher-generated CoT traces and monitoring outputs, omitting the generation and use of compact student explanations.
\textbf{StageGuard (Decision-Only)} removes CoT supervision and trains the student directly on ground-truth monitoring decisions and next-task labels.

\subsection{System Integration Test}
We evaluate StageGuard's impact on task success in closed-loop hierarchical control on BEHAVIOR-1K and two real robot platforms. A PDDL planner~\cite{Garrett2020-cr} generates subtask sequences, with pick-and-place handled by motion planning and other skills by task-specific \(\pi_{0.5}\) policies~\cite{intelligence2025}. StageGuard governs subtask progression and policy switching through \(d_t\) and \(v_t\).
We fine-tune \(\pi_{0.5}\) on the corresponding BEHAVIOR-1K training data and 50 demonstrations per subtask for real-robot deployment. The single-arm UR5e opens a drawer, transfers a plate from a rack into the cabinet, and closes the drawer. The Piper arm performs a more dexterous task, picking up a plate and inserting it into a rack.

\subsection{Evaluation metrics}
We report six evaluation metrics. \textbf{Transition completion} is the fraction of required transitions correctly detected within $\pm 3$ seconds of their ground-truth times. \textbf{Full trajectories} is the percentage of trajectories reaching their final planned subtask. \textbf{Next-subtask accuracy} measures agreement between predicted and intended next subtasks. \textbf{Inference speed} measures examples processed per second (Hz). For integrated-system evaluation, we report \textbf{task success rate}, the percentage of trials completing the overall task, and \textbf{progress score}, the fraction of required subgoals achieved.

\begin{table*}[t]
\centering
\caption{Stage-transition evaluation and inference efficiency on BEHAVIOR-1K.
The first three metrics are reported as percentages, and inference
efficiency is measured in Hz. Higher values are better for all metrics
($\uparrow$). The best and second-best distinct values are shown in
\textbf{bold} and \underline{underlined}, respectively; ties receive
the same formatting.}
\label{tab:stageguard_b1k}

\fontsize{11pt}{11pt}\selectfont
\setlength{\tabcolsep}{3pt}
\renewcommand{\arraystretch}{1.05}

\begin{tabular*}{\textwidth}{@{\extracolsep{\fill}}lcccc@{}}
\toprule
Method
& \shortstack{Transition\\completion}
& \shortstack{Full\\trajectories}
& \shortstack{Next-subtask\\accuracy}
& \shortstack{Inference\\speed} \\
\midrule

GPT-5.6 Luna
& 1.96 & 0.00 & 0.24 & 0.59 \\

Gemma 4 E4B
& 0.00 & 0.00 & 0.00 & 6.09 \\

Gemma 4 12B Unified
& 1.96 & 0.00 & 1.92 & 4.95 \\

Qwen3.5-4B
& 1.96 & 0.00 & 1.20 & 1.23 \\

Qwen3.5-0.8B
& 0.00 & 0.00 & 0.00 & 4.72 \\

\midrule

ROVER-4B
& 3.92 & 0.00 & 1.20 & 1.21 \\

ROVER-0.8B
& 0.00 & 0.00 & 0.00 & \underline{6.28} \\

\midrule

StageGuard (w/o explain)
& 0.00 & 0.00 & 0.00 & 1.42 \\

StageGuard (Decision-Only)
& \underline{37.72}
& \underline{15.00}
& \underline{11.42}
& \textbf{35.49} \\

\textbf{StageGuard (ours)}
& \textbf{90.18}
& \textbf{81.67}
& \textbf{57.23}
& 1.74 \\

\bottomrule
\end{tabular*}
\end{table*}

\section{Results}

In this section, we present evaluation results to address the following questions: (1) Can StageGuard learn accurate stage-transition decisions and next-task predictions? (2) How does StageGuard compare with other VLMs and progress reasoning methods? (3) How efficient is StageGuard at inference? (4) Does integrating StageGuard into hierarchical robot control improve task success? 

\paragraph{Learning stage-transition decisions.}
StageGuard consistently achieves the highest transition completion rate and next-subtask accuracies on both benchmarks as shown in \Cref{tab:stageguard_libero,tab:stageguard_b1k}. Transition completion rates reach 96.23\% on LIBERO and 90.18\% on BEHAVIOR-1K. Although accuracy decreases on the more challenging BEHAVIOR-1K tasks, performance remains strong despite unseen subtask decompositions and execution orders in the evaluation trajectories, which require generalization beyond the training sequences. Both results substantially exceed those of the underlying Qwen3.5-0.8B model. These improvements demonstrate that the proposed training pipeline enables a lightweight VLM to learn accurate stage-transition decisions and next-task predictions from demonstrations.

\paragraph{Comparison with baseline models and methods.}
We compare StageGuard with open-weight and proprietary VLMs, as well as VLM-based progress reasoning methods. The larger Gemma and Qwen baselines remain substantially below StageGuard performance, suggesting limitations in both their pretrained reasoning capabilities and their alignment with subtask completion criteria. GPT-5.6 Luna performs better than the open-weight baselines on LIBERO but struggles on BEHAVIOR-1K, suggesting that general-purpose reasoning does not automatically yield decisions aligned with subtask completion criteria and transition timing. ROVER-4B clearly improves upon its backbone on LIBERO, but its gains on BEHAVIOR-1K are minimal. This suggests that structuring inference-time reasoning can help with simpler tasks while remaining constrained by the underlying model's capabilities on more complex sequences. Together, these comparisons support demonstration-grounded reasoning distillation as an effective means of adapting lightweight VLMs to stage-transition monitoring.

\paragraph{Inference efficiency.}
As shown in \Cref{tab:stageguard_libero,tab:stageguard_b1k}, smaller pretrained VLMs generally offer lower latency, but their low transition accuracy limits the value of more frequent monitoring. GPT-5.6 Luna provides stronger predictions on LIBERO at substantially higher time cost, illustrating the accuracy-efficiency trade-off of relying on general pretrained models. StageGuard achieves the highest transition completion rate while sustaining inference throughputs of 1.03~Hz on LIBERO and 1.74~Hz on BEHAVIOR-1K. These results demonstrate improvements in both prediction accuracy and inference speed over GPT-5.6 Luna.

\subsection{Ablation Study}

For the ablation study, we mainly aim to answer Question (5): What are the contributions of CoT supervision and learning by explanation?

The full StageGuard model achieves the highest transition completion rate and next-subtask accuracies on both benchmarks, with a substantially larger advantage on more complex tasks. On LIBERO, teacher CoT traces are relatively short and simple, making direct imitation more manageable for the lightweight student. Decision-only supervision can also capture transition patterns within constrained, largely fixed scenes. Consequently, both variants approach the full StageGuard's transition completion rate.
The advantage becomes more pronounced on BEHAVIOR-1K, where longer mobile manipulation sequences introduce changing viewpoints, more varied task contexts, and unseen subtask orderings. Decision-only supervision provides no explicit explanation of how these observations and task dependencies determine completion, while directly imitating longer, more complex teacher traces may exceed what the lightweight student can effectively learn. Learning by explanation helps address both challenges by retaining decision-relevant reasoning in compact student-generated explanations. The larger gains on BEHAVIOR-1K therefore suggest that adapting the form of reasoning supervision becomes increasingly valuable as task complexity grows.

\subsection{System Integration}

As shown in \Cref{tab:stageguard_system}, StageGuard substantially improves closed-loop task success, bringing system performance closer to that achieved with ground-truth stage transitions. This highlights the importance of accurate transition monitoring: an incorrect switching decision can prematurely terminate a subtask or trigger an inappropriate next skill, allowing errors to propagate through subsequent stages. Consequently, decision-only monitoring provides only a modest improvement over the end-to-end policy, as hierarchical planning alone cannot ensure reliable execution when stage transitions are inaccurate. By improving transition decisions, StageGuard narrows the success-rate gap to the oracle from 0.28 to 0.10, enabling the system to realize more of the benefits of hierarchical control. The oracle's remaining failures further indicate that low-level execution remains a limiting factor even with correct stage transitions.

\begin{table}[t]
\centering
\caption{System integrated evaluation on BEHAVIOR-1K, measured by task success rate and progress score, both reported on a 0--1 scale. Higher values are better for both metrics ($\uparrow$). The best and second-best results are shown in \textbf{bold} and \underline{underlined}, respectively.}
\label{tab:stageguard_system}

\fontsize{10pt}{10pt}\selectfont
\setlength{\tabcolsep}{3pt}
\renewcommand{\arraystretch}{1.05}

\begin{tabularx}{\columnwidth}{@{}Xcc@{}}
\toprule
Method
& \shortstack{Success\\rate}
& \shortstack{Progress\\score} \\
\midrule

End-to-end learned policy
& 0.30 & 0.43 \\

\midrule
\multicolumn{3}{@{}l}{\textit{Hierarchical planning +}} \\

GT transitions (oracle)
& \textbf{0.63} & \textbf{0.88} \\

StageGuard (Decision-only)
& 0.35 & 0.54 \\

\textbf{StageGuard (ours)}
& \underline{0.53} & \underline{0.76} \\

\bottomrule
\end{tabularx}
\end{table}

\begin{figure*}[t]
    \centering
    \includegraphics[width=1.0\textwidth]{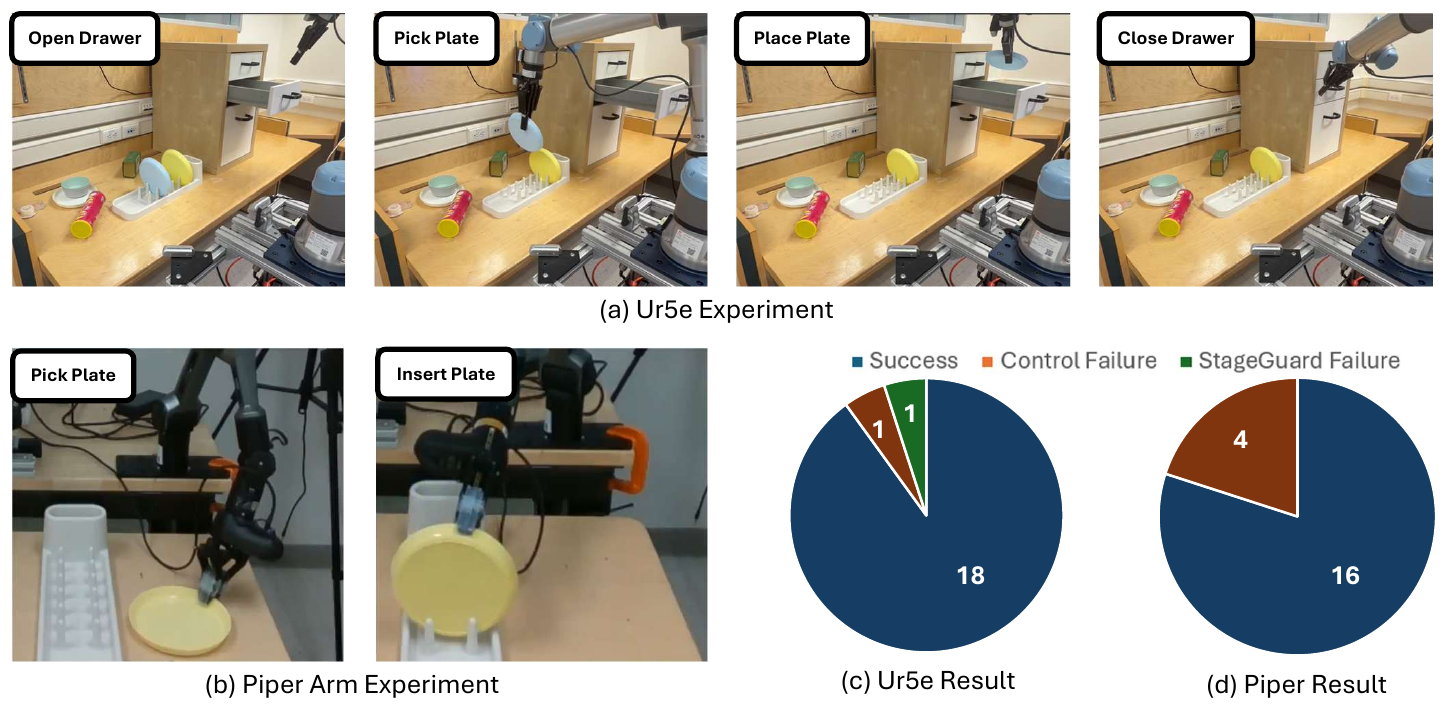}
    \caption{Real-robot evaluation of StageGuard. (a) Real robot experiment on UR5e. (b) Real robot experiment on Piper Arm. (c) Evaluation result on UR5e (d) Evaluation result on Piper Arm.
}
    \label{fig3}
    \label{fig:real_robot}
\end{figure*}

\section{Real-Robot Experiment}
\Cref{fig:real_robot} shows the UR5e and Piper arm experimental settings. We train StageGuard on the 50 collected demonstration trajectories and conduct 20 closed-loop trials in each setting, measuring integrated-system success rate and the monitoring latency introduced by StageGuard.
The single-arm system succeeds in 18 of 20 trials (90\%). One failure results from VLA control, while the other arises from a missed transition between plate placement and drawer closing due to partial observability. The piper system succeeds in 16 of 20 trials (80\%). All four failures result from unsuccessful grasps of the thin plate edge, with no failures attributed to StageGuard. The model runs at 2.2 Hz in parallel with the local VLA, without observable blocking of the control loop.

\section{Conclusion}

We presented StageGuard, an agentic distillation framework for accurate and efficient stage-transition monitoring. Its teacher workflow generates structured explanations linking observations and task requirements to demonstrated transitions. Learning by explanation transfers these decision patterns into a lightweight student through compact self-explanations. Evaluations on LIBERO and BEHAVIOR-1K demonstrate improved transition prediction and fast inference. Ablation studies further show that adapting teacher reasoning to the student is particularly beneficial on more complex mobile manipulation tasks. Real robot experiments on UR5e and piper arm systems demonstrate effective integration into closed-loop hierarchical control across embodiments. These findings highlight the value of aligning pretrained reasoning with task-specific completion criteria while reducing inference latency.
Despite improvements in accuracy and efficiency, a few limitations remain. StageGuard relies on annotated demonstrations and computationally expensive supervision generation, remains sensitive to partial observability, and cannot compensate for low-level control failures. Future work will investigate more data-efficient adaptation, lower-cost supervision generation, active perception, and recovery policies.

\bibliographystyle{IEEEtran}
\bibliography{refs}

\end{document}